\documentclass[letterpaper, 10 pt, conference]{ieeeconf}

\IEEEoverridecommandlockouts                              

\usepackage{graphicx}
\usepackage{balance}
\usepackage{tabularx}
\usepackage{multirow}
\usepackage{multicol}
\usepackage{booktabs}
\usepackage{caption}
\usepackage{siunitx}
\usepackage{algorithm}
\usepackage{algorithmic}

\usepackage{amssymb}
\usepackage{amsmath}
\usepackage{subcaption}
\usepackage{tablefootnote}
\usepackage{bbding}
\usepackage{threeparttable}
\usepackage{xcolor}
\usepackage{overpic}

\newcommand{\MTDstate}{\mathcal{S}_{t}^{(v)}}
\newcommand{\bl}[1]{{\textcolor{blue}{#1}}}
\newcommand{\ol}[1]{{\textcolor{olive}{#1}}}
\newcommand{\rl}[1]{{\textcolor{red}{#1}}}
\newcommand{\gl}[1]{{\textcolor{green}{#1}}}

\newcommand{\grl}[1]{{\textcolor{gray}{#1}}}

\makeatletter
\let\NAT@parse\undefined
\makeatother
\usepackage[numbers,sort&compress]{natbib} 

\usepackage[bookmarks=true]{hyperref}
\hypersetup{
    colorlinks=true, 
    urlcolor=black,
    citecolor=blue,
    linkcolor=red
}

\usepackage{cleveref}

\begin{document}

\title{\LARGE \bf \textit{MTD-Map}: Single-Stage Long-Term LiDAR Map Maintenance Framework via Mixture Transition Distribution}


\author{TaeYoung Kim$^{1}$, Gilhwan Kang$^{1}$, Tae Ihn Kim$^{1}$, Seungwon Song$^{1}$ and Hun Keon Ko$^{1, *}$%
\thanks{$^{1}$TaeYoung Kim, Gilhwan Kang, Tae Ihn Kim, Seungwon Song and Hunkeon Ko are with the Robotics Lab, Advanced Vehicle Platform Division, Hyundai Motor Company, Uiwang-si, Gyeonggi-do, Republic of Korea, 16082 {\tt\footnotesize \{tyoung96, gilhwan, taeihn, ssw.robotics, hoonhoon\}@hyundai.com}}%
\thanks{$^{*}$  Corresponding author}%
}


\maketitle

\begin{abstract}

While robust map maintenance has advanced significantly, existing studies have focused on specific tasks, especially dynamic object removal or change detection.
In this paper, we take a holistic view of the map maintenance problem and propose MTD-Map, a single-stage framework that handles both dynamic object removal and change detection without separate task-specific modules.
MTD-Map employs an explicit representation that compactly encodes the direction and duration of occupancy transitions through Mixture Transition Distribution (MTD) modeling.
We develop a recursive MTD formulation that encodes historical occupancy patterns into an augmented state to capture high-order temporal dependencies.
Furthermore, a stability-driven adaptive strategy balances noise suppression with the preservation of quasi-static structures.  
Extensive experiments verify that MTD-Map robustly removes dynamic objects and achieves competitive change detection performance, subsequently reducing computational costs.
Our project page is available at: \url{https://taeyoung96.github.io/mtd_map/}.

\end{abstract}

\vspace{8pt}

\section{INTRODUCTION}


Long-term mapping using light detection and ranging (LiDAR) sensors enables persistent robotic operations over extended periods \cite{tipaldi2013lifelong, lazaro2018efficient, zou2024ltaom}.
Real-world environments undergo spatiotemporal variations. Thus, pre-built maps inevitably become outdated, causing localization failures \cite{akai2022detection} and compromising safe navigation.
Effective map maintenance is vital for long-term reliability, yet it faces challenges in real-world environments.
Specifically, environments continuously evolve. They contain short-term dynamic obstacles (e.g., pedestrians), quasi-static entities (e.g., parked vehicles), and permanent structures.

\begin{figure}[t]
\centering
\begin{overpic}[width=1.0\columnwidth]{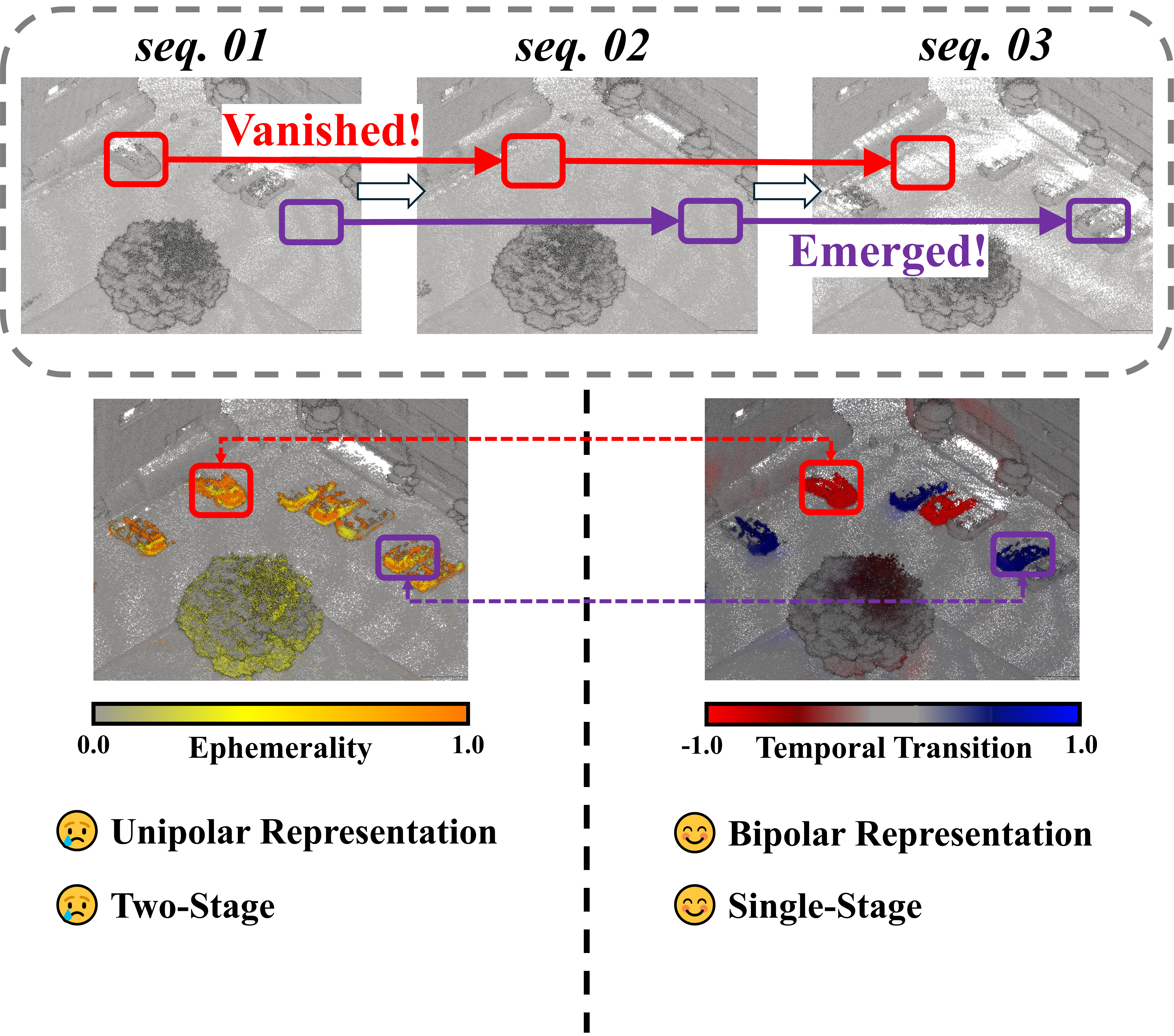}
\put(13, 2.5){\small (a) ELite \cite{gil2025elite}}
\put(60, 2.5){\small (b) MTD-Map (Ours)}
\end{overpic}
\caption{Illustration of map representations in the \textit{LT-ParkingLot} sequence. While ELite \cite{gil2025elite} relies on a directionless, unipolar representation, MTD-Map employs a bipolar representation to naturally distinguish between arriving and vacating events. This enables continuous modeling of quasi-static entities without incurring additional computational costs.}
\vspace{-0.7cm}
\label{fig1}
\end{figure}

Despite progress in map maintenance over the past decade, prior works have addressed specific tasks such as dynamic object removal (DOR) and change detection (CD).
DOR methods \cite{lim2021erasor, lim2023erasor2, duberg2024dufomap, wu2024otd, bhandari2025hmmmos} filter short-term dynamic entities within a single session.
In contrast, CD approaches \cite{wellhausen2017reliable, schmid2022panoptic, adam2022objects, rowell2024lista} identify spatial discrepancies across multiple sessions to update long-term maps.
Lifelong mapping frameworks, which continuously maintain a consistent static map over extended periods, have attempted to integrate both DOR and CD capabilities. 
However, capturing the continuous evolution of map elements remains challenging, often leading to information loss or high computational overhead.
Modular approaches \cite{zhao2021general, kim2022ltmapper, yang2024lifelong} enforce binary decisions, categorizing occupancy changes as either `appearing' or `vacating' events. 
These discretizations tend to ignore the probabilistic spectrum of persistence, causing information loss in environments with gradual changes.
A recent probabilistic approach, ELite \cite{gil2025elite} models the gradual evolution of map elements through a two-stage ephemerality scheme.
However, its reliance on directionless map representations causes directional ambiguity, as shown in \Cref{fig1}. This ambiguity necessitates additional computation to distinguish arrival and departure events for CD.



We holistically address the map maintenance problem by framing DOR and CD as unified inference of occupancy transitions over varying timescales.
DOR targets short-term, transient fluctuations, and CD focuses on long-term shifts in persistent states.
Bridging these timescales requires modeling a continuous persistence spectrum—from transient objects and quasi-static entities to permanent structures.
Hidden Markov Models \cite{wang2014iohmm} theoretically capture sequential transitions. 
However, first-order assumptions and static matrices restrict their ability to model the required continuum.
High-order Markov chains, in turn, capture this continuum but induce exponential complexity.
This prohibits their application to large-scale mapping.
Therefore, we adopt the Mixture Transition Distribution (MTD) \cite{berchtold2002mixture} to efficiently approximate high-order dependencies.
Specifically, MTD formulates transition probabilities as a linear combination of past states, thereby reducing the parameter space from exponential to linear.
MTD decomposes temporal inference into an immediate response and an aggregated history.
This bridges the timescales of DOR and CD, enabling a single-stage probabilistic update.


The main contributions of this paper are summarized as follows:

\begin{itemize}
 
\item We introduce MTD-Map, a unified probabilistic framework that integrates DOR and CD into a single stage. This streamlines map updates by bypassing modular redundancies.

\item We present a recursive MTD formulation utilizing an augmented voxel state to compactly encode transient and persistent spatial states, thereby enabling high-order temporal inference with bounded computational cost. 

\item We propose a stability-driven adaptive strategy that balances the suppression of dynamic objects and the preservation of static structures based on prediction fidelity.

\item We demonstrate our approach on real-world datasets, showing improved accuracy compared to existing methods.
Our framework reduces processing time relative to state-of-the-art (SOTA) methods that perform both DOR and CD.

\end{itemize}

The remainder of this paper is organized as follows:
\Cref{related-works} reviews existing map maintenance methods.
\Cref{proposed_framework} details the proposed MTD-Map framework.
\Cref{experiments} validates the proposed approach on real-world datasets.
Finally, \Cref{conclusions} concludes the paper with future directions.


\section{Related Works}
\label{related-works}

In general, map maintenance is dictated by two primary tasks: DOR addresses short-term dynamics, while CD addresses long-term structural changes.
Traditionally, these tasks have been addressed independently; however, recent research has unified them into a cohesive framework.

\subsection{Dynamic Object Removal}

DOR eliminates dynamic objects to construct clean static maps. \textit{Occupancy-based} methods such as OctoMap \cite{hornung2013octomap} and Dynablox \cite{schmid2023dynablox} clear dynamic observations via probabilistic ray tracing. DUFOMap \cite{duberg2024dufomap} identifies void regions using single-scan ray casting, while HMM-MOS \cite{bhandari2025hmmmos} segments moving points by tracking temporal voxel states via Hidden Markov Models. 
Alternatively, \textit{discrepancy-based} methods—Removert \cite{kim2020removert}, ERASOR \cite{lim2021erasor}, and OTD \cite{wu2024otd}—evaluate geometric mismatches against accumulated submaps using multi-resolution images, pseudo-occupancy ratios, and observation time differences, respectively.
\textit{Learning-based} methods such as SwiftMOS \cite{11358413} require supervised training to address per-scan moving object segmentation rather than persistent map maintenance.
However, DOR strictly targets short-term dynamic filtering and ignores long-term persistence, necessitating an additional task for holistic map maintenance.

\subsection{Change Detection}

Conversely, CD identifies long-term structural evolution. \citet{underwood2013explicit} employs spherical ray tracing to mitigate false positives caused by occlusions. \citet{wellhausen2017reliable} and \citet{fehr2017tsdf} detect modifications by analyzing residuals in Truncated Signed Distance Fields (TSDF) or utilizing real-time clustering for change identification. More recently, LiSTA \cite{rowell2024lista} adopts object-level tracking to monitor structural shifts, while Chamelion \cite{jang2026chamelion} leverages 4D sparse convolutional neural
network with synthetic augmentation to differentiate changes in transient environments. While effective at localizing changes, these approaches predominantly operate on a discrete session-to-session basis. Such discretizations capture the final result of a change rather than the underlying transition process, fundamentally limiting unified temporal reasoning.

\subsection{Temporal Representations for Lifelong Mapping}

Early works such as \citet{pomerleau2014longterm} addressed long-term map maintenance by removing dynamic elements without explicit object models.
LT-mapper \cite{kim2022ltmapper} and the method proposed by \citet{yang2024lifelong} integrate LiDAR-based mapping with DOR and CD but rely on sequential modules.
This modularity separates structural estimation and dynamic processing, precluding unified temporal inference.

Spatiotemporal approaches encode environmental evolution but remain limited by scalar history compression \cite{gil2025elite} or periodic assumptions \cite{krajnik2017fremen}.
Similarly, ProbPer-LiLo \cite{ali2026probperlilo} prioritizes the extraction of permanent structures via geometric properties rather than modeling explicit state transitions.
Unlike these approaches, our proposed method explicitly models both the duration and direction of occupancy transitions to disambiguate emerging and vanishing entities.



\section{MTD-MAP: Unified Temporal Framework \\ For Robust Map Maintenance}
\label{proposed_framework}

\begin{figure}[b]
\centering
\includegraphics[width=1\columnwidth]{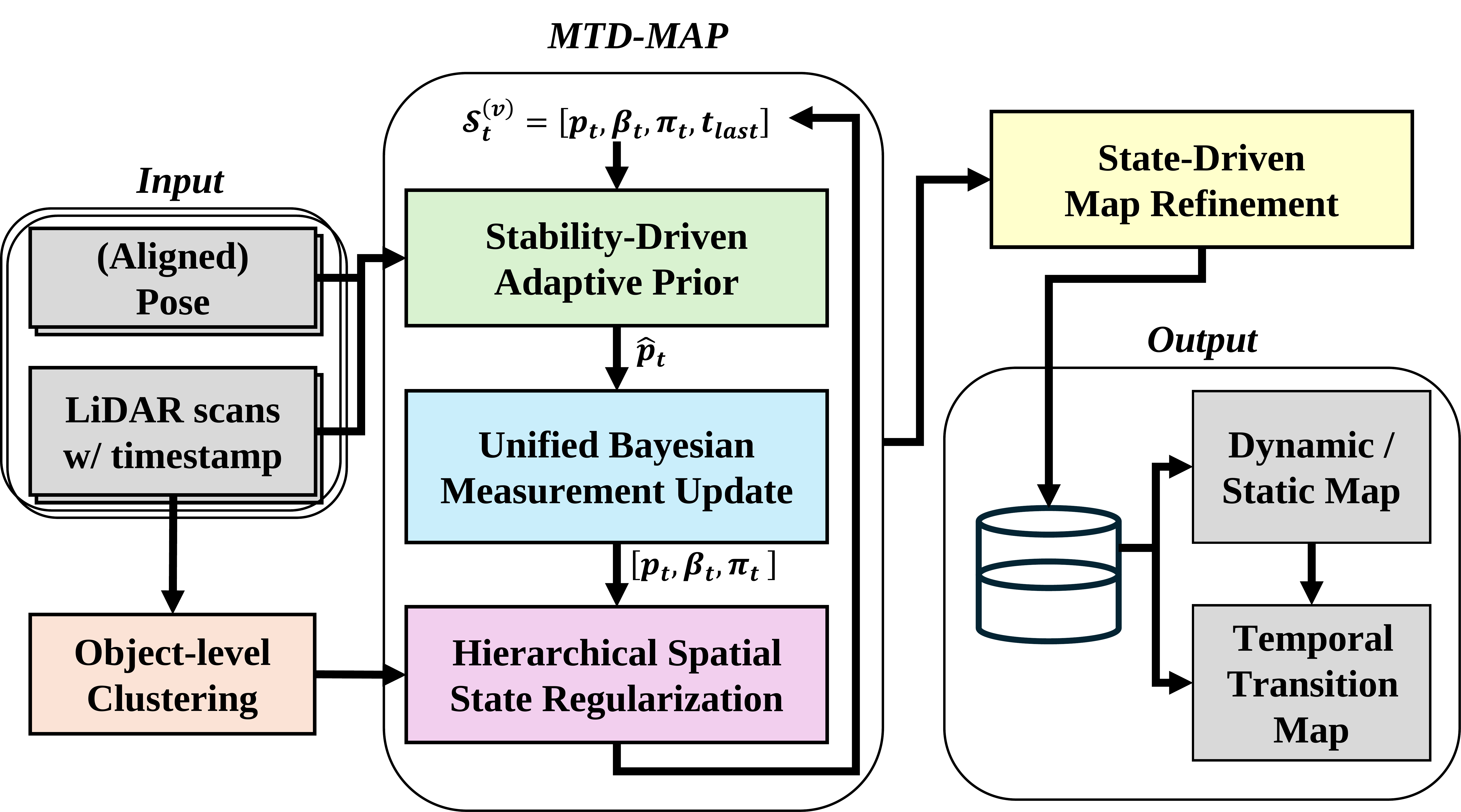}
\caption{Overview of our proposed MTD-Map framework.}
\vspace{-0.5cm}
\label{fig:overview}
\end{figure}

\subsection{System Overview}
\label{subsec:overview}

As illustrated in \Cref{fig:overview}, the proposed MTD-Map framework concurrently addresses DOR and CD within a unified single-stage architecture.
The system accepts a stream of globally aligned LiDAR scans, their respective timestamps, and the corresponding sensor poses as input.

We perform object-level clustering on each incoming frame as a preprocessing step for spatial regularization.
This spatial grouping feeds into the core logic, which employs MTD-based temporal modeling.
We maintain an augmented state $\MTDstate$ for each voxel $v$ to capture structural persistence.
Before integrating new measurements, a stability-driven adaptive prior calculates a time-decayed prediction.
This prior balances instantaneous occupancy and long-term history adaptively based on voxel stability and elapsed time.
Subsequently, the unified Bayesian measurement update incorporates the incoming scan into the augmented state vector.
This mechanism adjusts the stability based on the prediction fidelity to ensure adaptive responsiveness.

Following the update, a hierarchical spatial state regularization step enforces local belief consistency.
This regularization leverages the initial clusters to filter local outliers and propagate representative states.
Finally, a map refinement step translates the converged augmented states into two map representations.
The module first decouples the environment into static and dynamic maps.
Subsequently, it generates a temporal transition map to capture spatiotemporal state transitions.

\subsection{State Augmentation}
\label{state-augmentation}

Traditional occupancy grid mapping, such as OctoMap, assumes that the current occupancy state $\mathcal{O}_{t}$ depends solely on the predecessor $\mathcal{O}_{t-1}$:
\begin{equation}
P(\mathcal{O}_{t} | \mathcal{O}_{t-1}, z_{1:t}) = P(\mathcal{O}_{t} | \mathcal{O}_{t-1}, z_{t})
\end{equation}
where $z_{t}$ denotes the sensor observation at time $t$.
This assumption simplifies computation, but lacks the historical patterns required to distinguish dynamic objects from quasi-static and static states.

To address this issue, we define an augmented state vector for each voxel $v$:
\begin{equation}
\mathcal{S}_{t}^{(v)} = [p_t, \pi_{t}, \beta_{t}, t_{last}]
\end{equation}
where $p_{t}$ and $\pi_{t}$ represent the instantaneous and long-term occupancy probabilities, respectively. The term $\beta_{t}$ quantifies the temporal stability, which is mapped to a range $[0, 1]$ via the sigmoid function $\sigma(\beta_{t}) = (1 + e^{-\beta_{t}})^{-1}$ during updates.
Storing the most recent update time of the individual voxel $t_{last}$ allows the framework to explicitly model state evolution during unobserved intervals. Finally, we initialize the augmented state vector as follows: $\mathcal{S}_{0}^{(v)}=[0.5, \, 0.5, \, 0.0, \, 0.0]$.

\subsection{High-Order Temporal Modeling via MTD}
\label{subsec:mtd}

We model state transitions using the MTD, which approximates high-order Markov dependencies through a linear mixture of first-order transitions.
A standard $k$-order MTD sums over previous $k$ lags:
\begin{equation}
P(\mathcal{S}_{t} \mid \mathcal{S}_{t-1}, \dots, \mathcal{S}_{t-k}) = \sum_{j=1}^{k} \lambda_{j} \cdot q(\mathcal{S}_{t} \mid \mathcal{S}_{t-j})
\label{eq:mtd-equation}
\end{equation}
where $k$ is the Markov order, $q(\cdot)$ is the transition kernel, and $\lambda_j$ are the mixing weights satisfying $\sum_{j=1}^{k} \lambda_j = 1$.
Although this formulation captures long-term dependencies, storing $k$ discrete states per voxel is computationally prohibitive for large-scale mapping.

To ensure efficiency independent of history length $k$, we reformulate \cref{eq:mtd-equation} by partitioning it into an immediate response ($j=1$) and an aggregate historical component ($j \geq 2$).
Crucially, instead of retaining the explicit sequence $\mathcal{S}_{t-2}, \dots, \mathcal{S}_{t-k}$, we summarize their aggregate influence with the long-term occupancy probability $\pi_{t-1}$, which serves as a sufficient statistic for the history. This substitution yields a recursive form with constant memory cost:
\begin{equation}
\begin{split}
P(\mathcal{S}_{t} \mid \mathcal{S}_{t-k}^{t-1}) ={} & \lambda_{1} \cdot q(\mathcal{S}_{t} \mid \mathcal{S}_{t-1}) \\
& + (1 - \lambda_{1}) \cdot q(\mathcal{S}_{t} \mid \pi_{t-1})
\end{split}
\label{eq:partitioned_mtd}
\end{equation}
Consequently, \cref{eq:partitioned_mtd} expresses the transition as a linear mixture of two explanatory variables, the instantaneous state $\mathcal{S}_{t-1}$ and the historical summary $\pi_{t-1}$, an equivalence grounded in MTD regression theory~\cite{berchtold2002mixture}.


\subsection{Stability-Driven Adaptive Prior}
\label{subsec:adaptive-prior}

We derive a recursive update rule from the MTD formulation by substituting the static coefficient $\lambda_1$ with a time-adaptive weight $w_t$:
\begin{equation}
    \hat{p}_t = w_t \cdot p_{t-1} + (1 - w_t) \cdot \pi_{t-1}
    \label{eq:recursive_update}
\end{equation}
Specifically, $w_t$ balances the instantaneous occupancy probability $p_{t-1}$ and the long-term occupancy probability $\pi_{t-1}$.

To implement \cref{eq:recursive_update}, we first define the effective time constant $\tau_{e}$ as a function of stability $\beta_{t}$:
\begin{equation} 
    \tau_{e}(\beta_{t}) = \tau_{init} \cdot \left( 1 + \gamma \cdot \sigma(\beta_{t}) \right)
    \label{eq:tau_eff} 
\end{equation} 
Here, $\tau_{init}$ and $\gamma$ govern the nominal decay rate and temporal scaling, respectively.
The stability $\sigma(\beta_{t})$ modulates responsiveness: high values increase $\tau_{e}$ to preserve established structures, while low values allow rapid adaptation.

Using \cref{eq:tau_eff}, we formulate the weight $w_{t}$ as a stability-modulated decay factor based on the elapsed time $\Delta t = t_{current} - t_{last}$:
\begin{equation}
    w_t = \left( \frac{1}{1 + \frac{\Delta t}{\tau_{e}(\beta_{t-1})}} \right)^{2}
    \label{eq:power_law_full}
\end{equation}
For substantial temporal gaps (e.g., session changes), the prior is adaptively selected based on long-term occupancy probability: $\pi_{t-1}$ is prioritized for high occupancy to ensure structural persistence, whereas $p_{t-1}$ takes precedence to prevent over-commitment to obsolete structures.


\subsection{Unified Bayesian Measurement Update}
\label{subsec:bayesian-update}

Unlike traditional binary Bayes filters that treat state updates in isolation, we model the state $\MTDstate$ of each voxel $v$ as a coupled system.
Within our framework, prediction fidelity drives the update of stability $\beta_t$, which subsequently modulates the balance between $p_t$ and $\pi_t$.
This coupling enables adaptive responsiveness to environmental changes.

\subsubsection{Instantaneous Occupancy Probability}
Given a LiDAR observation with timestamp $z_t$, we compute the posterior instantaneous occupancy probability $p_t$ using Bayes' rule:
\begin{equation} 
p_t = \frac{p(z_t \mid \mathcal{S}_t=1) \cdot \hat{p}_t}{p(z_t \mid \mathcal{S}_t=1) \cdot \hat{p}_t + p(z_t \mid \mathcal{S}_t=0) \cdot (1 - \hat{p}_t)} 
\label{eq:prob_update}
\end{equation}
where $p(z_t \mid \mathcal{S}_t)$ denotes the measurement likelihood.
We apply the logit transform $L(p) = \log(p / (1-p))$ to ensure numerical stability, converting the operation into a linear summation in log-odds space:
\begin{equation}
L(p_t) = L(\hat{p}_{t}) + L(p(z_{t} \mid \mathcal{S}_{t}))
\label{eq:logit_transform}
\end{equation}

\subsubsection{Long-term Occupancy Probability}

We update the long-term occupancy probability $\pi_{t}$ via an adaptive exponential moving average with a stability-modulated rate $\alpha_t$:
\begin{equation}
\pi_{t} = (1 - \alpha_t) \cdot \pi_{t-1} + \alpha_t \cdot p_{t}
\label{eq:pi_update}
\end{equation}
where the adaptive learning rate $\alpha_t$ is defined based on the base rate $\alpha_0$:
\begin{equation}
\alpha_t = \alpha_0 \cdot \sigma(\beta_{t-1})
\label{eq:alpha_rate}
\end{equation}
This recursive update is grounded in the infinite-lag MTD models \cite{berchtold2002mixture}, employing geometric lag weights equivalent to an exponential moving average. 
Consequently, $\pi_t$ provides a parsimonious summary of all previous states.

\subsubsection{Stability}
Stability $\beta_t$ evaluates state conservation by measuring the alignment between $z_t$ and $\hat{p}_t$ via discrepancy $\delta_t = |z_t - \hat{p}_t|$ for both `hit' and `miss' observations.
\begin{equation}
\beta_t =
\begin{cases}
\beta_{t-1} + \eta_{\text{inc}} \cdot (1 - \delta_{t}), & \delta_{t} < \epsilon \\
\kappa \cdot \beta_{t-1}, & \delta_{t} \geq \epsilon \land \sigma(\beta_{t-1}) \geq  \Gamma_{\beta} \\
\beta_{t-1} - \eta_{\text{dec}}, & \text{otherwise}
\end{cases}
\label{eq:stability_update}
\end{equation}
Consistency within tolerance $\epsilon$ reinforces $\beta_t$ to reward persistence, while conflicts exceeding $ \Gamma_\beta$ trigger a scaling penalty $\kappa$ for rapid adaptation. Other cases undergo linear decay $\eta_{dec}$ to mitigate transient noise.

The complete recursive inference procedure is summarized in \Cref{alg:unified_update}.

\begin{table*}[t!]
\centering
\caption{Quantitative evaluation of DOR in point cloud maps.}
\label{table:dynamic_object_removal}
\resizebox{\textwidth}{!}{%
\begin{tabular}{c|ccc|ccc|ccc|ccc}
\hline \hline
              & \multicolumn{3}{c|}{\textit{KITTI} \texttt{00}}                 & \multicolumn{3}{c|}{\textit{KITTI} \texttt{07}}             & \multicolumn{3}{c|}{\textit{HeLiMOS} \texttt{6593}}  &
              \multicolumn{3}{c}{\textit{MOE} \texttt{02}}
              \\ \hline\hline
Method        & SA$\uparrow$ & DA$\uparrow$  & HA$\uparrow$ & SA$\uparrow$ & DA$\uparrow$  & HA$\uparrow$ & SA$\uparrow$  & DA$\uparrow$ & HA$\uparrow$ & SA$\uparrow$  & DA$\uparrow$ & HA$\uparrow$      \\ \hline
Removert \cite{kim2020removert} & 99.08 & 68.53 & 81.02 & 99.12 & 34.06 & 50.70 & 97.42 & 40.65 & 57.37 & 96.78 & 27.20 & 42.46 \\
ERASOR \cite{lim2021erasor} & 90.44 & \textbf{96.37} & 93.31 & 89.19 & \textbf{97.37} & \underline{93.10} & 82.58 & \underline{95.61} & 88.62 & 94.55 & \underline{99.38} & \textbf{96.90} \\
DUFOMap \cite{duberg2024dufomap} & 99.19 & 83.46 & 90.64 & \underline{99.14} & 55.60 & 74.25 & 98.08 & 67.64 & 80.07 & \underline{96.42} & 48.13 & 64.21 \\
OTD \cite{wu2024otd} & 94.93 & 93.74 & \underline{94.33} & 94.64 & 84.54 & 89.31 & 97.93 & 73.27 & 83.83 & 95.67 & 87.50 & 91.40 \\
HMM-MOS \cite{bhandari2025hmmmos} & \textbf{99.92} & 58.45 & 73.76 & \textbf{99.82} & 45.57 & 67.44 & \textbf{99.56} & 65.62 & 79.10 & \textbf{99.24} & 5.84 & 11.04 \\
ELite \cite{gil2025elite} & 92.09 & 89.24 & 90.64 & 93.77 & 90.27 & 91.99 & 89.28 & \textbf{95.80} & \textbf{92.43} & 82.93 & \textbf{99.76} & 90.57 \\
MTD-Map (Ours) & \underline{99.47} & \underline{95.28} & \textbf{97.33} & 92.71 & \underline{95.61} & \textbf{94.13} & \underline{98.51} & {82.95} & \underline{90.06} & 95.95 & {94.84} & \underline{95.39} \\
\hline
\end{tabular}%
}
\begin{minipage}{18.5cm}
\vspace{0.1cm}
*The best and second best results are shown in \textbf{bold} and \underline{underline}, respectively. The units are in \%.
\end{minipage}
\vspace{-0.5cm}
\end{table*}

\subsection{Hierarchical Spatial State Regularization}
\label{subsec:regularization}

We apply spatial regularization at each time step $t$ to mitigate occlusion and ensure local belief consistency. 
Leveraging TRAVEL~\cite{oh2022travel}, we partition the scene into a set of geometrically coherent clusters $\{\mathcal{C}_l\}$, where $l \in \{1, \dots, N\}$ denotes the index of $N$ clusters. 
For each voxel $v \in \mathcal{C}_l$, we construct a feature vector $\mathbf{f}_t = {[p_t, \pi_t, \sigma(\beta_t)]}^\top$ derived from its augmented state $\MTDstate$. 
To reject outliers within each cluster, we compute the Mahalanobis distance relative to the cluster median. 
This metric naturally accounts for varying scales and inter-variable correlations among the state components:
\begin{equation}
    d_{\mathcal{M}}(\mathbf{f}_t) = \sqrt{(\mathbf{f}_t - \tilde{\mathbf{f}}^{\mathcal{C}_l})^\top \cdot (\boldsymbol{\Sigma}^{\mathcal{C}_l})^{-1} \cdot (\mathbf{f}_t - \tilde{\mathbf{f}}^{\mathcal{C}_l})},
    \label{eq:mahalanobis}
\end{equation}
where $\tilde{\mathbf{f}}^{\mathcal{C}_l}$ and $\boldsymbol{\Sigma}^{\mathcal{C}_l}$ represent the median feature vector and the covariance matrix of all voxels within cluster $\mathcal{C}_l$, respectively. 
Voxels satisfying $d_{\mathcal{M}} \leq \tau_{\mathcal{M}}$ are classified as mainstream, where the threshold $\tau_{\mathcal{M}}=3.0$ is empirically chosen to bound structural variance rather than act as a strict statistical confidence interval. 
Finally, we compute a representative feature $\mathbf{f}^{\mathcal{C}_l*}$ from these mainstream voxels and propagate it to update the global state, thereby reinforcing spatial consistency.

\begin{algorithm}[t!]
\caption{Recursive Augmented State Inference}
\label{alg:unified_update}
\begin{algorithmic}[1]
    \REQUIRE $z_t$, $t_{current}$, $\mathcal{S}_{t-1}^{(v)}$ \text{or} $\mathcal{S}_{0}^{(v)}$
    \ENSURE $\mathcal{S}_{t}^{(v)}$

    \STATE \textbf{// Stability-Driven Adaptive Prior}
    
    \STATE $\hat{p}_t \leftarrow \texttt{AdaptivePrior}(\mathcal{S}_{t-1}^{(v)} \text{ or } \mathcal{S}_{0}^{(v)}, t_{current})$ \\ \hfill $\triangleright$ eq. (\ref{eq:recursive_update},  \ref{eq:tau_eff}, \ref{eq:power_law_full})

    \STATE \textbf{// Unified Bayesian Measurement Update}
    
    \STATE $p_t \leftarrow \texttt{InstOccProbUpdate}(z_{t}, \hat{p}_{t})$ \hfill $\triangleright$ eq. (\ref{eq:prob_update}, \ref{eq:logit_transform})
    
    \STATE $\pi_{t} \leftarrow \texttt{LongTermOccProbUpdate}(p_{t}, \pi_{t-1}, \beta_{t-1})$ \\ \hfill $\triangleright$ eq. (\ref{eq:pi_update}, \ref{eq:alpha_rate})
    
    \STATE $\beta_t \leftarrow \texttt{StabilityUpdate}(z_{t}, \hat{p}_{t}, \beta_{t-1})$ \hfill $\triangleright$ eq. (\ref{eq:stability_update})
    
    \STATE $t_{last} \leftarrow t_{current}$ \textbf{// Time Update}
    
    \STATE $\mathcal{S}_{t}^{(v)} \leftarrow [p_t, \pi_t, \beta_t, t_{last}]$  \textbf{// State Replacement}
    
    \RETURN $\mathcal{S}_{t}^{(v)}$
\end{algorithmic}
\end{algorithm}

\subsection{State-Driven Map Refinement}

Complementing \Cref{subsec:regularization}, we refine $\mathcal{S}^{(v)}_{sub} = [p, \pi, \beta]$ for viewpoint-invariant global geometric coherence.
We define the static set via the threshold vector $\mathcal{T}_{sta} = [\tau_p^{sta}, \tau_\pi^{sta}, \tau_\beta^{sta}]$:
\begin{equation}
  \mathcal{V}_{static} = \{v \mid p > \tau_p^{sta}, \, \pi > \tau_\pi^{sta}, \, \beta < \tau_\beta^{sta}\}
  \label{eq:dynamic_th}
\end{equation}
Voxels failing these conditions form the dynamic set $\mathcal{V}_{dynamic} = \mathcal{V} \setminus \mathcal{V}_{static}$, removed in DOR. We then smooth $\mathcal{V}_{static}$ with a trilateral filter over spatial proximity, surface normals, and state similarity, preserving geometric features.
  

Next, we partition $\mathcal{V}_{static}$ into three subsets to generate a temporal transition map $\Psi$. The appeared ($\mathcal{V}_{AQS}$, currently occupied) and disappeared ($\mathcal{V}_{DQS}$, currently vacated) quasi-static voxel sets are defined by threshold vectors $\mathcal{T}_{aqs}$ and $\mathcal{T}_{dqs}$:
\begin{equation}
\begin{aligned}
    \mathcal{V}_{AQS} &= \{v \mid p > \tau_{p}^{aqs}, \, \pi > \tau_{\pi}^{aqs}, \, \beta < \tau_{\beta}^{aqs} \}, \\
    \mathcal{V}_{DQS} &= \{v \mid p < \tau_{p}^{dqs}, \, \pi < \tau_{\pi}^{dqs}, \, \beta > \tau_{\beta}^{dqs} \},
\end{aligned}
\label{eq:quasi_static_sets}
\end{equation}
and the co-existing static voxel set is defined as the remainder: $\mathcal{V}_{CS} = \mathcal{V}_{static} \setminus (\mathcal{V}_{AQS} \cup \mathcal{V}_{DQS})$. 
In \cref{eq:bipolar_mapping}, $s^{(v)} \in [0, 1]$ represents the normalized final stability $\beta$ of voxel $v$.

\begin{equation}
\Psi(v) = \frac{1}{2}
\begin{cases}
\phantom{-}(1+s^{(v)}) & \text{if } v \in \mathcal{V}_{AQS} \\
\pm (1 - s^{(v)}) & \text{if } v \in \mathcal{V}_{CS} \\
-(1 + s^{(v)}) & \text{if } v \in \mathcal{V}_{DQS}
\end{cases}
\label{eq:bipolar_mapping}
\end{equation}
This projects transitions onto a unified $[-1, 1]$ spectrum, where the sign gives the change direction ($+$ for $\mathcal{V}_{AQS}$, $-$ for $\mathcal{V}_{DQS}$) and the magnitude distinguishes persistent transitions from stable structures.



\section{Experiments}
\label{experiments}

The proposed framework is comprehensively validated through both quantitative and qualitative evaluations across various public, real-world datasets. All experiments were conducted on a desktop, equipped with an Intel Xeon W-2245 CPU (8 cores, 16 threads, 3.90\,GHz) and 64\,GB of RAM. \Cref{tab:parameters} lists the user-defined hyperparameters used in our experiments. 
Detailed evaluation results regarding DOR and CD are presented in the following subsections.

\subsection{Dynamic Object Removal}
\begin{figure}[t]
\centering
\includegraphics[width=1\columnwidth]{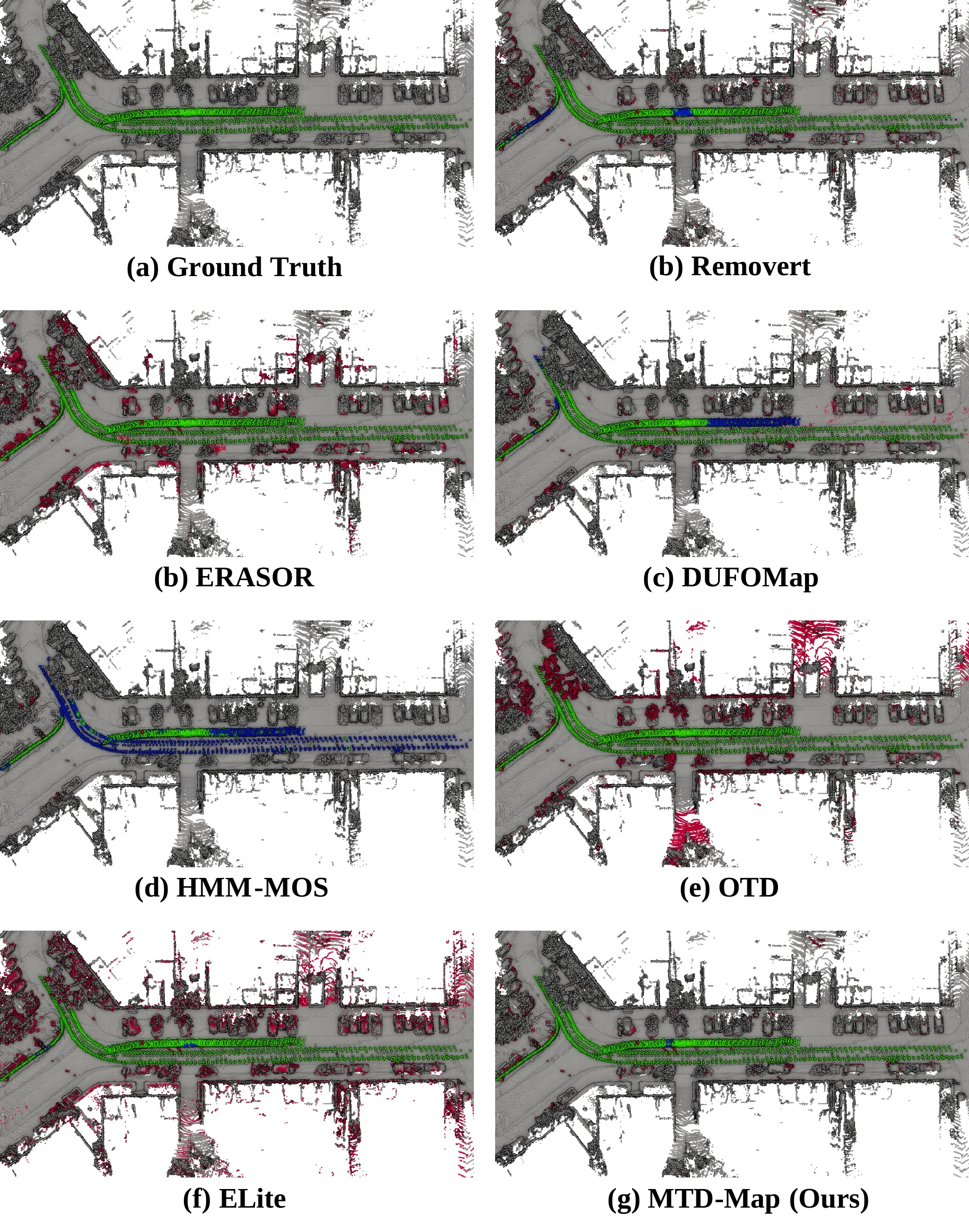}
\caption{Qualitative DOR results for \textit{SemanticKITTI} (seq. \texttt{00}) Points are visualized based on classification accuracy: True Positive (\textbf{\gl{green}}), False Negative (\textbf{\bl{blue}}), and False Positive (\textbf{\rl{red}}).
}
\label{fig3}
\vspace{-0.5cm}
\end{figure}

To evaluate the performance of DOR, we utilize \textit{SemanticKITTI} \cite{behley2019semantickitti}, \textit{HeLiMOS} \cite{lim2024helimos}, and \textit{MOE} \cite{chen2024moe} datasets, all of which provide high-fidelity, point-wise ground truth labels for dynamic objects.
As suggested in \cite{zhang2023dynamic} and \cite{jia2024beautymap}, we employ Static Accuracy (SA), Dynamic Accuracy (DA), and Harmonic Accuracy (HA) as our evaluation metrics, which quantify map-level maintenance quality rather than per-scan segmentation accuracy.
We evaluate all methods at a consistent voxel leaf size of $0.1\,\text{m}$ to isolate algorithmic differences from resolution effects.
We utilized the poses provided by the dataset to maintain consistency across all experiments.
MTD-Map is compared against SOTA methodologies, including Removert, ERASOR, DUFOMap, HMM-MOS, OTD, and ELite. 
Learning-based methods require per-sensor supervised training, precluding a fair comparison with our training-free framework.
The comparative results are detailed in \Cref{table:dynamic_object_removal} and visualized in \Cref{fig3}.

\begin{table}[t!]
\centering
\caption{Parameters of our proposed method.}
\label{tab:parameters}
\begin{tabular}{@{}llp{4.5cm}@{}}
\toprule
\textbf{Parameters} & \textbf{Value} & \textbf{Description} \\ \midrule
$v_{leaf}$          & 0.1 & Voxel leaf size \\
$\tau_{init}$       & 60.0 & Initial time constant for decay \\
$\gamma$            &  10.0      & Temporal scaling factor  \\
$\alpha_0$          & 0.02             & Base learning rate for $\pi$ \\
$\epsilon$          & 0.45             & Prediction error tolerance threshold \\
$\eta_{inc}$        & 1.0               & Stability increment rate  \\
$\eta_{dec}$        & 0.2              & Stability linear decay rate \\
$\kappa$            & 0.7          & Scaling penalty factor \\
$\Gamma_{\beta}$    & 0.8          & Stability threshold for rapid adaptation \\ \midrule
$\mathcal{T}_{sta}$    & $[0.3, \, 0.5, \, 5.0]$    & Threshold vector for $\mathcal{V}_{static}$ \\
$\mathcal{T}_{aqs}$    & $[0.49, \, 0.06, \, 90.0]$          & Threshold vector for $\mathcal{V}_{AQS}$ \\
$\mathcal{T}_{dqs}$    & $[0.2, \, 0.11, \, 96.0]$          & Threshold vector for $\mathcal{V}_{DQS}$ \\
\bottomrule
\end{tabular}
\vspace{-0.3cm}
\end{table}

The experimental results reveal a persistent trade-off in existing DOR methodologies.
HMM-MOS and DUFOMap exhibit a conservative bias, maintaining high SA but showing limited DA. This tendency leaves substantial ghost trail effects in the final static map.
Conversely, ERASOR and ELite adopt an aggressive strategy. While this maximizes DA, it leads to an overall degradation of SA, often discarding vital static landmarks  for navigation, as highlighted in \Cref{fig3}.

MTD-Map effectively bridges this gap by balancing static preservation and dynamic removal.
For DOR, voxels failing the static threshold $\mathcal{T}_{sta}$ are classified as the dynamic set $\mathcal{V}_{dynamic}$ and removed.
As shown in \Cref{table:dynamic_object_removal}, MTD-Map achieved the highest HA in \textit{KITTI} \texttt{00} and \texttt{07}, outperforming competing methods that typically excel in only a single metric.
While ERASOR and ELite exhibit higher peak DA in specific highly dynamic scenarios like \textit{HeLiMOS} and \textit{MOE}, they often suffer from over-cleansing the erroneous removal of static structures.
In contrast, MTD-Map employs a stability-driven adaptive strategy to ensure greater consistency across diverse environments, overcoming the dependency of fixed-threshold methods. 
These results confirm that our approach maintains map integrity, whereas conventional methods often sacrifice fidelity for higher dynamic removal rates.

\subsection{Change Detection}

To ensure robust validation across diverse scenarios, we evaluate our framework in both indoor and outdoor environments. 
For long-term structural CD, we utilize the outdoor \textit{LT-ParkingLot} dataset, with local odometry estimated via FAST-LIO2 \cite{xu2022fastlio2} and global alignment ensured by Uni-Mapper \cite{kang2025unimapper}. 
All methods receive identical globally aligned inputs, isolating CD accuracy from registration error.
Since \textit{LT-ParkingLot} lacks binary labels for environmental changes, we manually annotated changing points in regions densely populated with quasi-static objects to evaluate quantitative CD performance.

\begin{table}[t]
\centering
\caption{Quantitative CD results for \textit{LT-ParkingLot} (seq. \texttt{03} $\to$ \texttt{04}).}
\label{table:quantitative_comparison}
\setlength{\tabcolsep}{4.5pt} 
\begin{tabular}{lcccccc}
\toprule
\multirow{2}{*}{Method} & \multicolumn{3}{c}{PD} & \multicolumn{3}{c}{ND} \\
\cmidrule(lr){2-4} \cmidrule(lr){5-7} 
 & Prec. & Rec. & F1 & Prec. & Rec. & F1 \\
\midrule
LT-mapper \cite{kim2022ltmapper} & 0.094 & 0.636 & 0.164 & 0.745 & 0.625 & 0.680 \\
ELite \cite{gil2025elite} & 0.209 & 0.369 & 0.267 & \textbf{0.806} & 0.423 & 0.555 \\
\midrule
MTD-Map (w/o Reg.) & 0.107 & 0.423 & 0.107 & 0.695 & 0.808 & 0.747 \\
MTD-Map (w/ Reg.) & \textbf{0.618} & \textbf{0.670} & \textbf{0.643} & 0.685 & \textbf{0.842} & \textbf{0.755} \\
\bottomrule
\end{tabular}
\begin{minipage}{8.5cm}
\vspace{0.1cm}
*MTD-Map (Ours) of both without and with regularization are compared.
*The best results are highlighted in \textbf{bold}.
\end{minipage}
\vspace{-0.5cm}
\end{table}

Following the terminology of LT-mapper, we evaluate Positive Differences (PD) for appeared structures and Negative Differences (ND) for disappeared structures. 
We quantify performance across both categories using precision, recall, and the F1-score.
To benchmark our approach against the current SOTA lifelong mapping frameworks, we conduct comparative experiments with LT-mapper and ELite.

\begin{figure}[b]
\centering
\includegraphics[width=1.0\columnwidth, trim=2 1 1 2, clip]{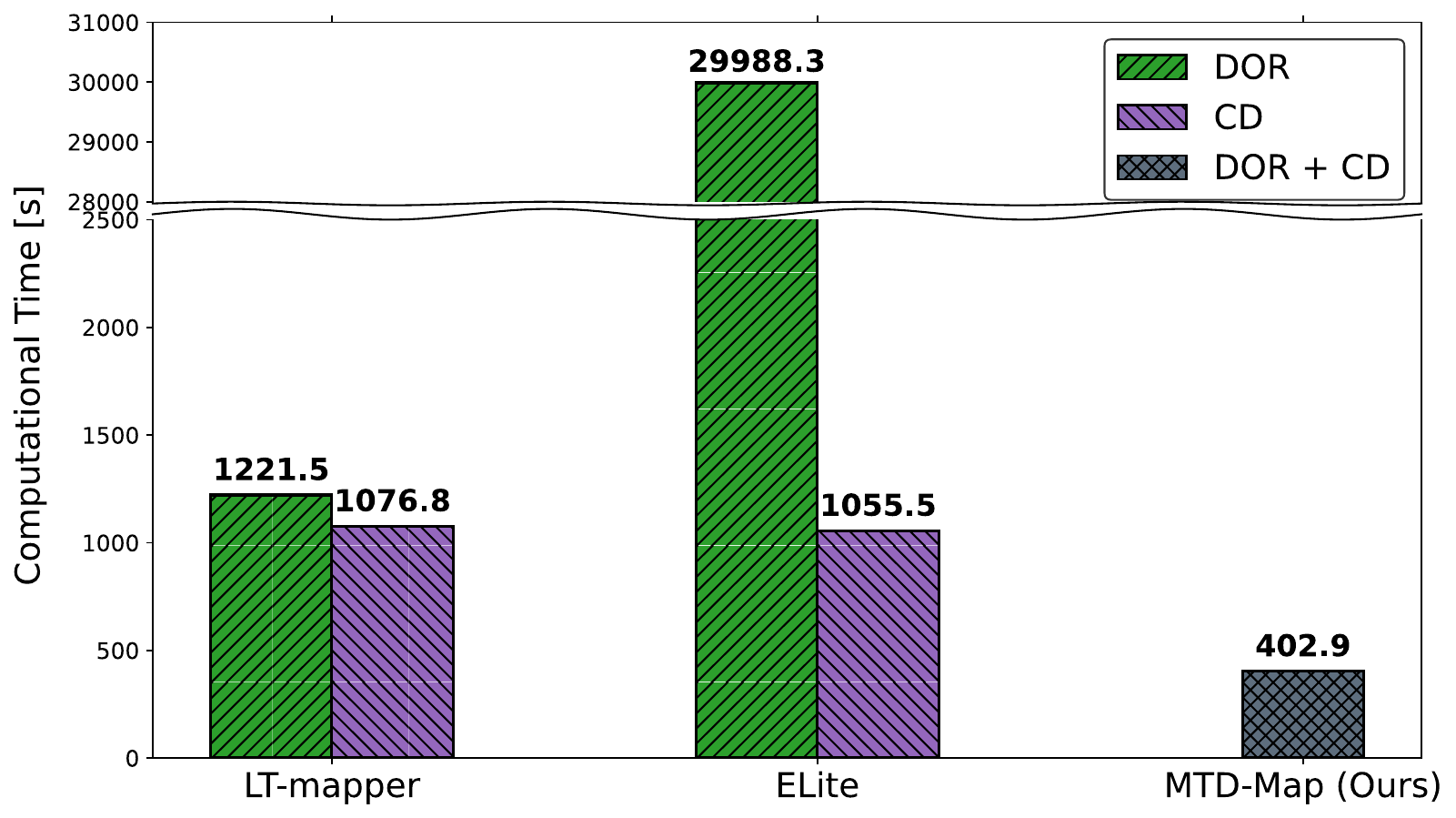}
\caption{Computation time comparison of each method using \textit{LT-ParkingLot} (seq. \texttt{03} $\to$ \texttt{04}).}
\vspace{-0.09cm}
\label{fig4}
\end{figure}

\begin{figure}[t]
\centering
\includegraphics[width=1.0\columnwidth]{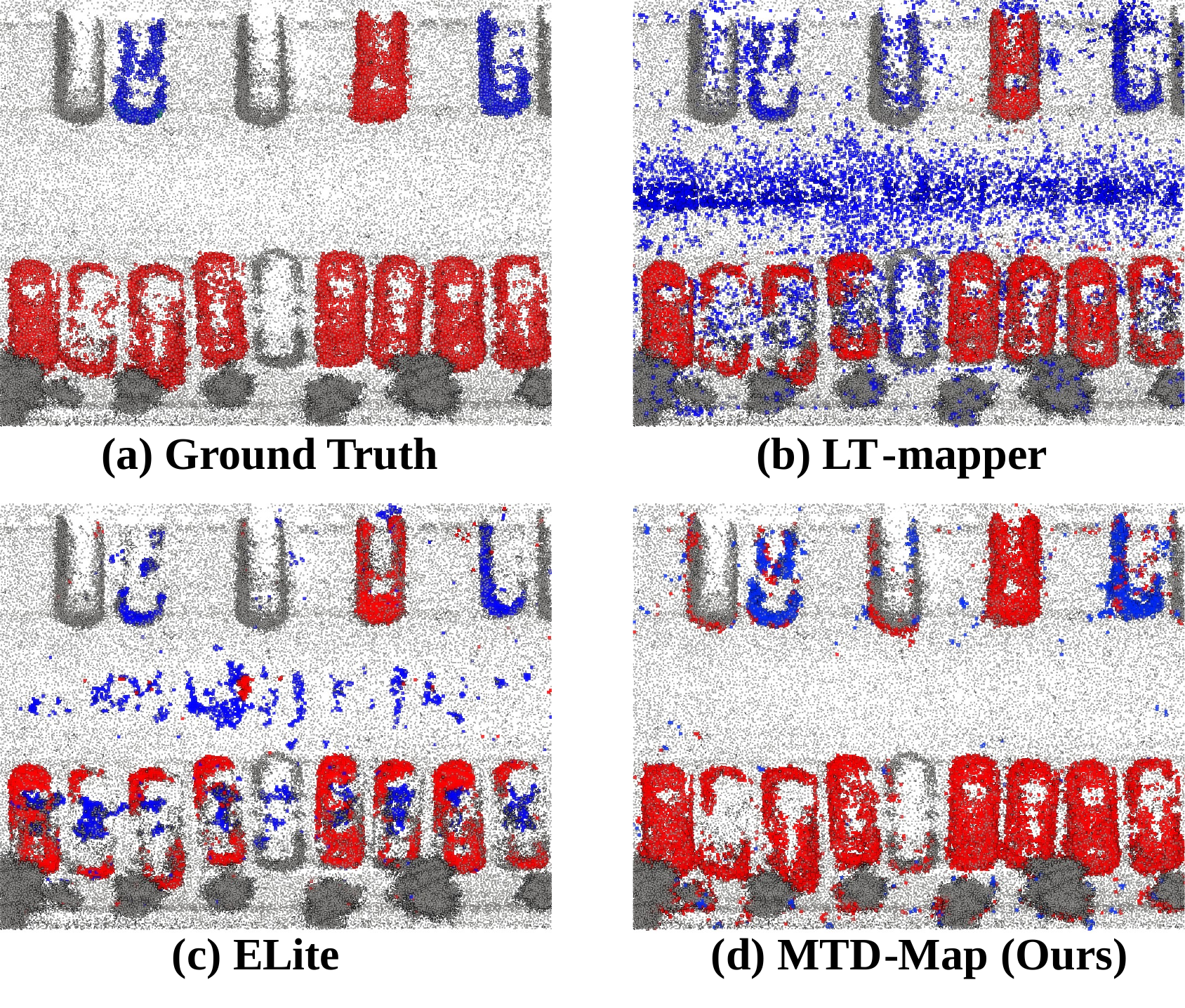}
\caption{
Qualitative CD results for \textit{LT-ParkingLot} (seq. \texttt{03} $\to$ \texttt{04}) highlighting PD (\textbf{\bl{blue}}) and ND (\textbf{\rl{red}}) points.
}
\vspace{-0.5cm}
\label{fig5}
\end{figure}

\Cref{fig5} and \Cref{table:quantitative_comparison} compare the proposed method against baselines on \textit{LT-ParkingLot} sequences \texttt{03} and \texttt{04}.
We selected these sequences for their balanced mixture of positive and negative changes.
LT-mapper suffers from high false positives on ground surfaces due to viewpoint-dependent visibility. While ELite attempts to mitigate these errors with a probabilistic model, its reliance on nearest-neighbor proximity leaves precision bottlenecks in PD detection unresolved.
Furthermore, while baselines achieve competitive precision in detecting ND points, their suboptimal recall indicates that the changed regions are only partially updated, leaving outdated artifacts.
This is particularly evident in ELite, which frequently misses the central portions of vehicles due to its dependency on an iterative point-wise update scheme, where low point density prevents the system from recognizing the full extent of the structural changes.

\begin{figure}[b]
\centering
\includegraphics[width=0.7\columnwidth]{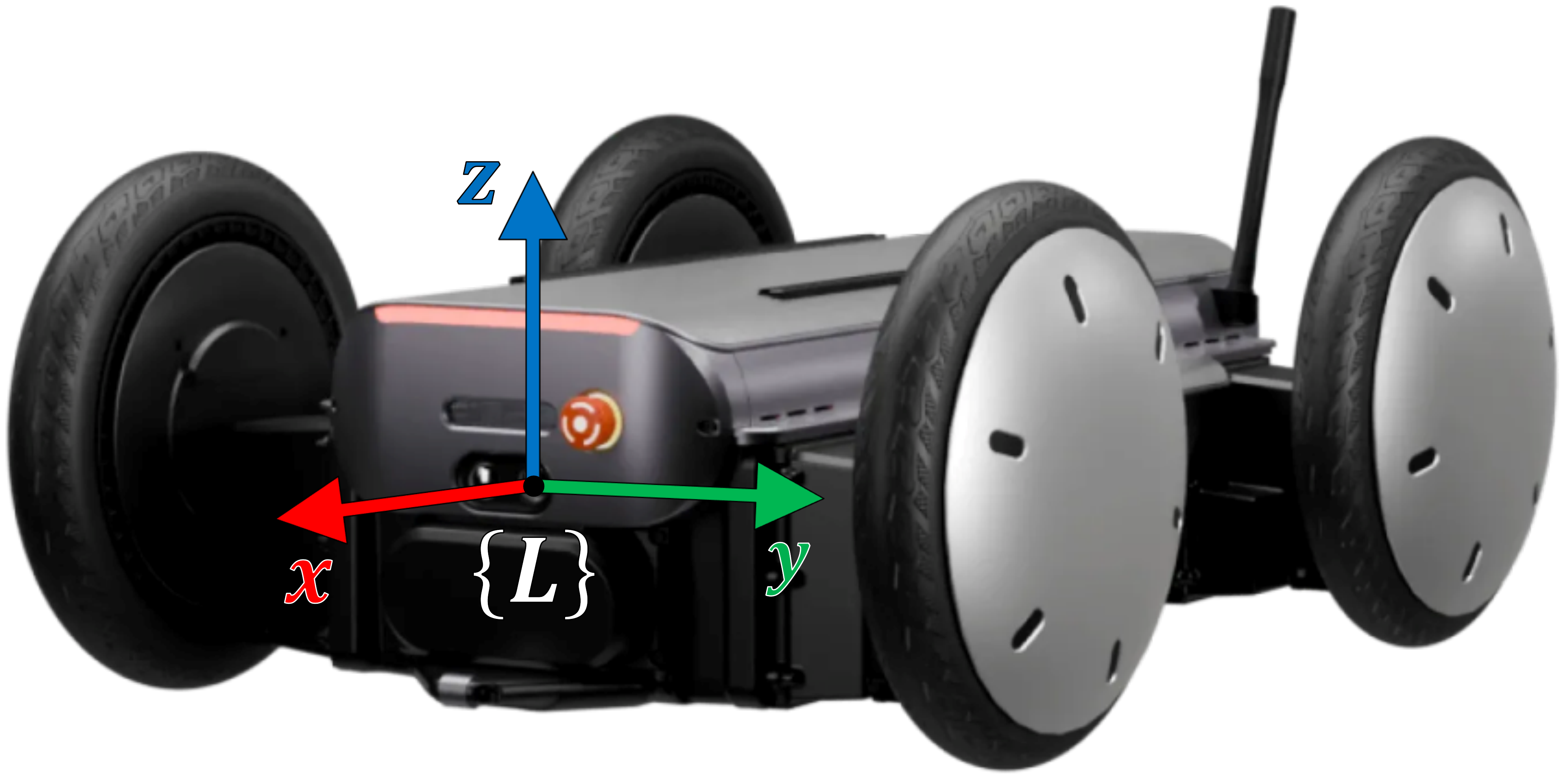}
\caption{The MobED platform used for custom data collection, where \textit{\{L\}} denotes LiDAR coordinate frame.}
\label{fig:mobed}
\vspace{-0.1cm}
\end{figure}

\begin{figure*}[t]
\centering
\includegraphics[width=2.05\columnwidth, trim=100 10 10 10, clip]{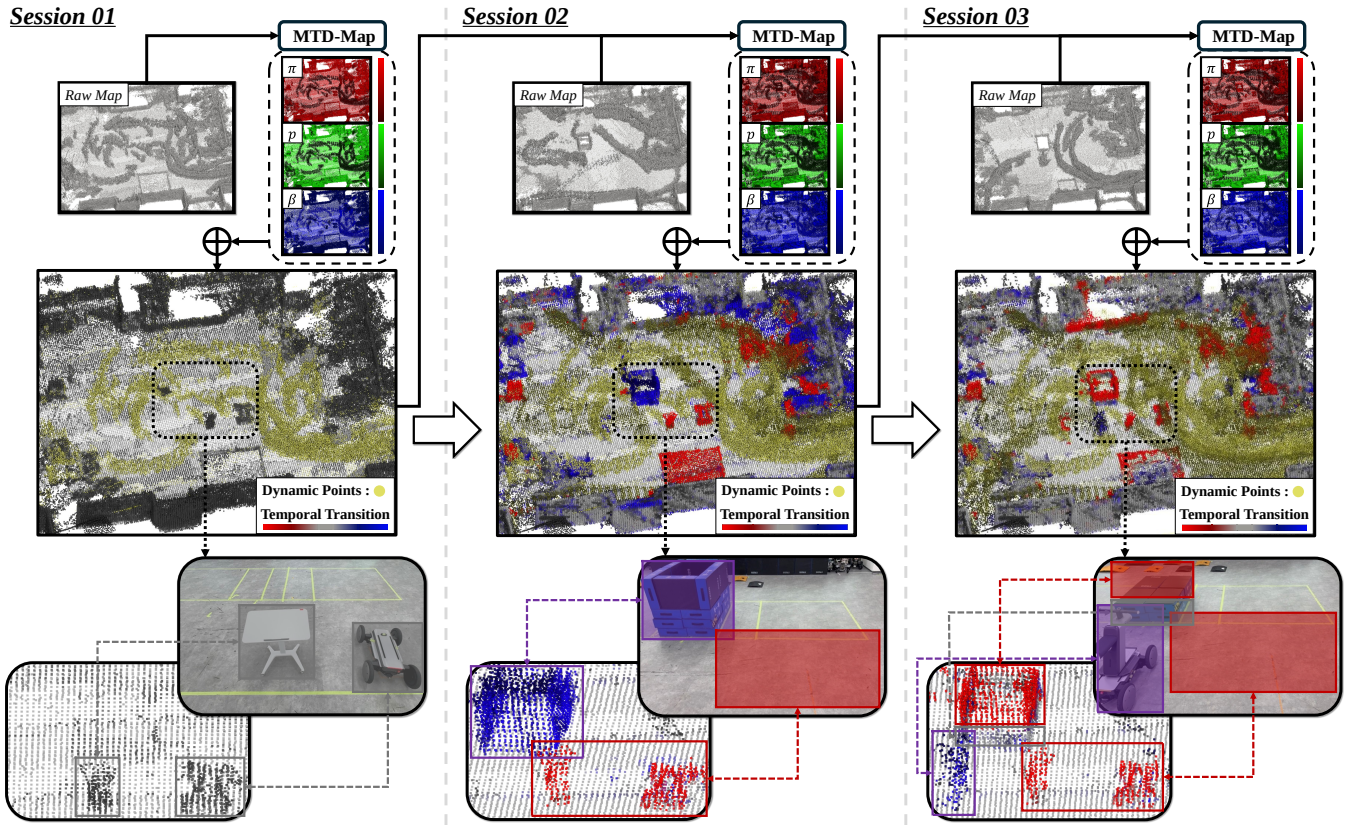}
\caption{
Qualitative DOR and CD results of the single-stage MTD-Map on the indoor dataset. \textbf{Top}: Raw point cloud map and corresponding MTD-based representations visualized by long-term occupancy probability ($\pi$, \textbf{\rl{red}}), instantaneous occupancy probability ($p$, \textbf{\gl{green}}), and stability ($\beta$, \textbf{\bl{blue}}) components.
\textbf{Center}: Temporal transition map $\Psi$ highlighting the dynamic points (\textbf{\ol{olive}}), visualizing the bipolar representation of environmental changes.
\textbf{Bottom}: Close-up visualizations of quasi-static objects, represented in \textbf{\grl{gray}}, alongside their corresponding camera views. In the point cloud maps, $\mathcal{V}_{AQS}$ and $\mathcal{V}_{DQS}$ points are represented in \textbf{\bl{blue}} and \textbf{\rl{red}}, respectively.
}

\label{main_fig}
\vspace{-0.35cm}
\end{figure*}

In contrast, MTD-Map achieves a superior F1-score via voxel-wise update strategy.
Our augmented state enables more sophisticated discrimination of temporal transitions compared to conventional binary or unipolar models.
To leverage this capability, we configure the joint thresholds as $\mathcal{T}_{aqs}$ and $\mathcal{T}_{dqs}$, delineating the regions of appearance and disappearance.
As evidenced by the ablation study in \Cref{table:quantitative_comparison}, 
this approach with probabilistic state regularization substantially improves PD precision. 
Specifically, this regularization mechanism ensures the robust identification of newly emerged structures while reliably reducing false positive detections in complex, semi-static environments.

To demonstrate the scalability of our approach in large-scale environments, we evaluate the computational efficiency of the proposed framework.
Conventional approaches typically separate DOR and CD into sequential tasks, incurring heavy computational costs due to repeated pointwise operations. 
In contrast, MTD-Map integrates these processes into a single voxel-wise update step.  
Consequently, as shown in \Cref{fig4}, our framework processes the entire pipeline in only 402.9 s, requiring approximately 25\% and 8\% of the total execution time needed by LT-mapper and ELite, respectively.
This unified architecture demonstrates substantial computational efficiency, demonstrating its suitability for long-term mapping in large-scale environments.

To further validate versatility, experiments were conducted in dynamic indoor environments using the Mobile Eccentric Droid (MobED) \cite{ko2026mobed} platform. As illustrated in \Cref{fig:mobed}, this platform is equipped with dual 3D solid-state LiDAR sensors and a single IMU.
\Cref{main_fig} illustrates the temporal map updates and identifies the areas of change over three consecutive sessions in a custom indoor environment.
Our augmented state inherently encodes both instantaneous dynamic obstacles and structural transitions within the representation itself. 
Dynamic objects exhibit low values across all components, whereas quasi-static changes are encoded into distinct bipolar transitions based on their temporal state profiles.
By leveraging these nuanced state distributions, MTD-Map effectively filters measurement artifacts and viewpoint-dependent noise, maintaining map consistency while precisely capturing the emergence and disappearance of objects. 
These results confirm that our stability-driven adaptive strategy, which preserves voxel states unless explicitly invalidated, delivers robust performance not only in outdoor parking scenarios but also across highly dynamic and complex indoor settings.



\section{CONCLUSION}
\label{conclusions}

We present MTD-Map, a unified probabilistic framework for dynamic object removal and change detection in long-term LiDAR map maintenance.
Our single-stage architecture distinguishes dynamic objects and captures structural evolution by integrating high-order temporal modeling with hierarchical spatial regularization, removing the need for separate modules.
The current framework assumes sequential sensor poses. 
Future work will focus on extending the framework to handle out-of-order pose inputs for real-time deployment. 
Furthermore, we aim to integrate MTD-Map into long-term autonomous navigation systems.


\footnotesize
\bibliographystyle{packages/IEEEtranN}  
\bibliography{bib}

\end{document}